\documentclass{article}

\usepackage[preprint]{neurips_2026}

\usepackage[utf8]{inputenc} 
\usepackage[T1]{fontenc}    
\usepackage{hyperref}       
\usepackage{url}            
\usepackage{booktabs}       
\usepackage{amsfonts}       
\usepackage{nicefrac}       
\usepackage{microtype}      
\usepackage{xcolor}         
\usepackage{amsmath}
\usepackage{graphicx}
\usepackage{algorithm}
\usepackage{algpseudocode}
\usepackage{subcaption}

\setcitestyle{round}

\title{Hybrid-LoRA: Bridging Full Fine-Tuning and Low-Rank Adaptation for Post-Training}

\author{%
  Chengqian Zhang$^{1}$ \quad Wei Zhu$^{2}$ \quad Kyumin Lee$^{1}$\\
  $^{1}$Worcester Polytechnic Institute \quad $^{2}$University of Hong Kong \\
  \texttt{czhang12@wpi.edu} \quad
  \texttt{wzhu91@connect.hku.hk} \quad
  \texttt{kmlee@wpi.edu}
}

\begin{document}

\maketitle

\begin{abstract}
Post-training has become essential for adapting large language models (LLMs) to complex downstream behaviors, including instruction following, preference alignment, and multi-step reasoning. Reinforcement learning with verifiable rewards (RLVR) has recently emerged as a particularly effective post-training paradigm for improving reasoning capabilities, with critic-free algorithms such as GRPO and GSPO enabling scalable optimization. However, RLVR post-training with full fine-tuning (FFT) requires substantial GPU memory and incurs high training costs. Although parameter-efficient fine-tuning (PEFT) methods, such as Low-Rank Adaptation (LoRA), effectively reduce computational costs, they often suffer from a noticeable performance gap compared to full fine-tuning in post-training for complex reasoning tasks. In this paper, we propose Hybrid-LoRA, an efficient hybrid post-training framework that selectively applies full fine-tuning to a small subset of modules less suited to low-rank adaptation, while adapting the remaining components with LoRA. We introduce a novel Hybrid-LoRA Score to rank candidate modules according to their sensitivity to low-rank adaptation under a fixed parameter budget. Experiments show that Hybrid-LoRA closely matches full fine-tuning performance under a 10\% full fine-tuning module budget, with the remaining candidate modules adapted by LoRA, consistently outperforming four state-of-the-art PEFT post-training baselines, achieving improvements of up to 5.65\% and on average 4.36\% over the best baseline.
\end{abstract}

\section{Introduction}
\label{sec: introduction}
Large language models (LLMs) have demonstrated remarkable proficiency in complex reasoning tasks through post-training, including supervised fine-tuning (SFT) and reinforcement learning (RL)-based approaches such as reinforcement learning from human feedback (RLHF) and reinforcement learning with verifiable rewards (RLVR)~\citep{ouyang2022,shao2024deepseek}. Recent critic-free RLVR algorithms, such as Group Relative Policy Optimization (GRPO) and Group Sequence Policy Optimization (GSPO), have become widely used for reasoning-oriented post-training, as they optimize policies using verifiable rewards without requiring a separate critic model. These methods have shown strong performance across downstream tasks, including mathematical reasoning, code generation, and scientific question answering. In the post-training stage of general-purpose large language models, existing RLHF and RLVR pipelines commonly rely on full fine-tuning (FFT), where all model parameters are updated, to adapt pre-trained or mid-trained foundation models for alignment and downstream tasks \citep{ouyang2022}. However, FFT requires substantial computational resources and is often bottlenecked by GPU memory and training cost.
Although parameter-efficient fine-tuning (PEFT) methods such as LoRA substantially reduce the cost of adapting large foundation models~\citep{houlsby19,hu2022lora}, their low-rank update space can limit model expressiveness, leading to a performance gap compared to full fine-tuning.

Several studies have sought to improve the expressiveness and parameter allocation of LoRA-based adaptation. For example, AdaLoRA \citep{zhang2023adaptive} decomposes the LoRA update into singular-value components and dynamically allocates the parameter budget by pruning less important components during training. AdaLoRA treats every singular value in the middle matrix with its two corresponding vectors as a triplet, computes a importance score for each of the triplets by multiplying its sensitivity and uncertain scores, and then prunes the ranks with lower importance scores based on the computation budget. Another approach, AutoLoRA \citep{zhang2024autolora} propose a bi-level optimization method to find the important rank among evenly distributed ranks in LoRA. AutoLoRA provides a contribution weight for each rank, and then update these weights based on the loss evaluated on validation set after every training step. The high contributing ranks (with higher weights) will be preserved and the model will be retrained when the ranks selection weights converge. Both approaches improve LoRA by allocating its limited rank budget more effectively across modules. However, they still operate entirely within the low-rank adaptation space, leaving a persistent gap from full fine-tuning.

With these insights in mind, we aim to bridge this gap by breaking the binary opposition between full fine-tuning and LoRA: rather than choosing one exclusively, we develop a hybrid fine-tuning framework that combines full-parameter updates with low-rank adaption. Our key hypothesis is that Transformer decoder modules are not equally amenable to low-rank adaptation. Some modules require full-parameter updates to preserve post-training performance, whereas others can be effectively adapted with LoRA. This idea introduces two major technical problems: (1) How can we efficiently estimate each module's suitability for low-rank adaptation and allocate full fine-tuning and PEFT under a fixed parameter budget? (2) How to achieve better performance of RL post-training in the hybrid fine-tuning framework with limited trainable parameters? To address these challenges, we develop a unified framework that integrates suitability estimation with stable RL optimization.

In this paper, our contributions are threefold:
\begin{itemize}
    \item We propose \textbf{Hybrid-LoRA}, a hybrid parameter optimization framework for LLM post-training that selectively assigns Transformer modules to FFT while adapting the remaining modules with LoRA. 
    \item We introduce \textbf{Hybrid-LoRA Score}, an efficient module-level evaluation metric based on first-order gradient-weighted sensitivity, to estimate each module's suitability for low-rank adaptation.
    \item We evaluate our framework across six complex reasoning tasks and various scales, demonstrating its effectiveness on a wide range of downstream benchmarks, including mathematical reasoning, scientific question answering, and code generation tasks. It outperforms four state-of-the-art PEFT baselines, achieving improvements of up to 5.65\% and on average 4.36\% relative improvement over the strongest baseline.   
\end{itemize}

\section{Related Works}
\label{sec: related_works}

\paragraph{Large Language Model Post-Training.} The evolution of LLM post-training has progressed from supervised fine-tuning (SFT) to reinforcement learning from human feedback (RLHF), and more recently to reinforcement learning with verifiable rewards (RLVR). Proximal Policy Optimization (PPO)~\citep{schulman2017proximal} has been widely used in early RLHF pipelines, where an additional value model is trained to estimate token-level value functions and compute advantages. However, maintaining this critic introduces substantial memory and computational overhead during large-scale LLM post-training. To reduce this overhead, recent RLVR methods have adopted critic-free optimization. \citet{shao2024deepseek} propose Group Relative Policy Optimization (GRPO), which samples multiple responses for each prompt and normalizes rewards within the group to estimate relative advantages without a separate critic. \citet{yu2025dapo} introduce DAPO, which improves large-scale long-CoT RL training with decoupled clipping, dynamic sampling, and other system-level and algorithmic refinements. \citet{zheng2025group} propose Group Sequence Policy Optimization (GSPO), which replaces token-level importance ratios with sequence-level importance ratios and performs sequence-level clipping to improve training stability.

\paragraph{Parameter-Efficient Fine-Tuning and LoRA Variants.} To reduce the substantial computational and memory burden of full-parameter adaptation, parameter-efficient fine-tuning (PEFT) methods, particularly LoRA \citep{hu2022lora} and its variants, have emerged as a practical alternative for adapting large language models. Several LoRA variants improve efficiency by automatically reallocating the limited rank budget across modules. AdaLoRA~\citep{zhang2023adaptive} introduces a dynamic rank allocation framework that parameterizes LoRA updates with singular-value components and progressively prunes less sensitive components based on a sensitivity-based importance score. SoRA and AutoLoRA ~\citep{ding2023sparse,zhang2024autolora} are closely related in their parameterization, as both decompose LoRA updates into sums of rank-one components formed by paired down- and up-projection vectors. SoRA introduces learnable gates over these rank-one components and optimizes them jointly with the LoRA parameters using proximal gradient updates under $L_0$ regularization, driving some gates exactly to zero so that the corresponding rank components can be pruned. AutoLoRA adopts a bi-level meta-learning framework, where continuous selection variables over rank-one components are optimized according to validation performance, and discrete layer-wise ranks are obtained by thresholding these variables. Different from the approaches above, where LoRA ranks are set to maximum and pruned to meet the target budget, ALoRA \citep{liu2024alora} initializes the rank with target budget uniformly distributed to all modules, calculates importance scores for each rank and reassigns pruned ranks to the modules where no ranks are pruned. 

All the above approaches aim to improve LoRA by reallocating its limited rank budget across modules or rank components. However, they remain restricted to the low-rank adaptation space, and thus cannot address cases where the required task-specific update for a module is not well captured by a low-rank parameterization. This limitation is particularly relevant in RLVR post-training, where complex reasoning behaviors may require richer updates in certain Transformer modules. Therefore, rather than asking only how to allocate LoRA capacity, we ask which modules are sufficiently handled by LoRA and which modules require full-rank adaptation. Based on this perspective, we propose Hybrid-LoRA, a hybrid adaptation framework that heterogeneously assigns full fine-tuning or low-rank adaptation to different modules according their suitability for LoRA adaptation.

\section{The Proposed Hybrid-LoRA Framework}
\label{sec: methodology}
\begin{figure}
    \centering
    \includegraphics[width=1\linewidth]{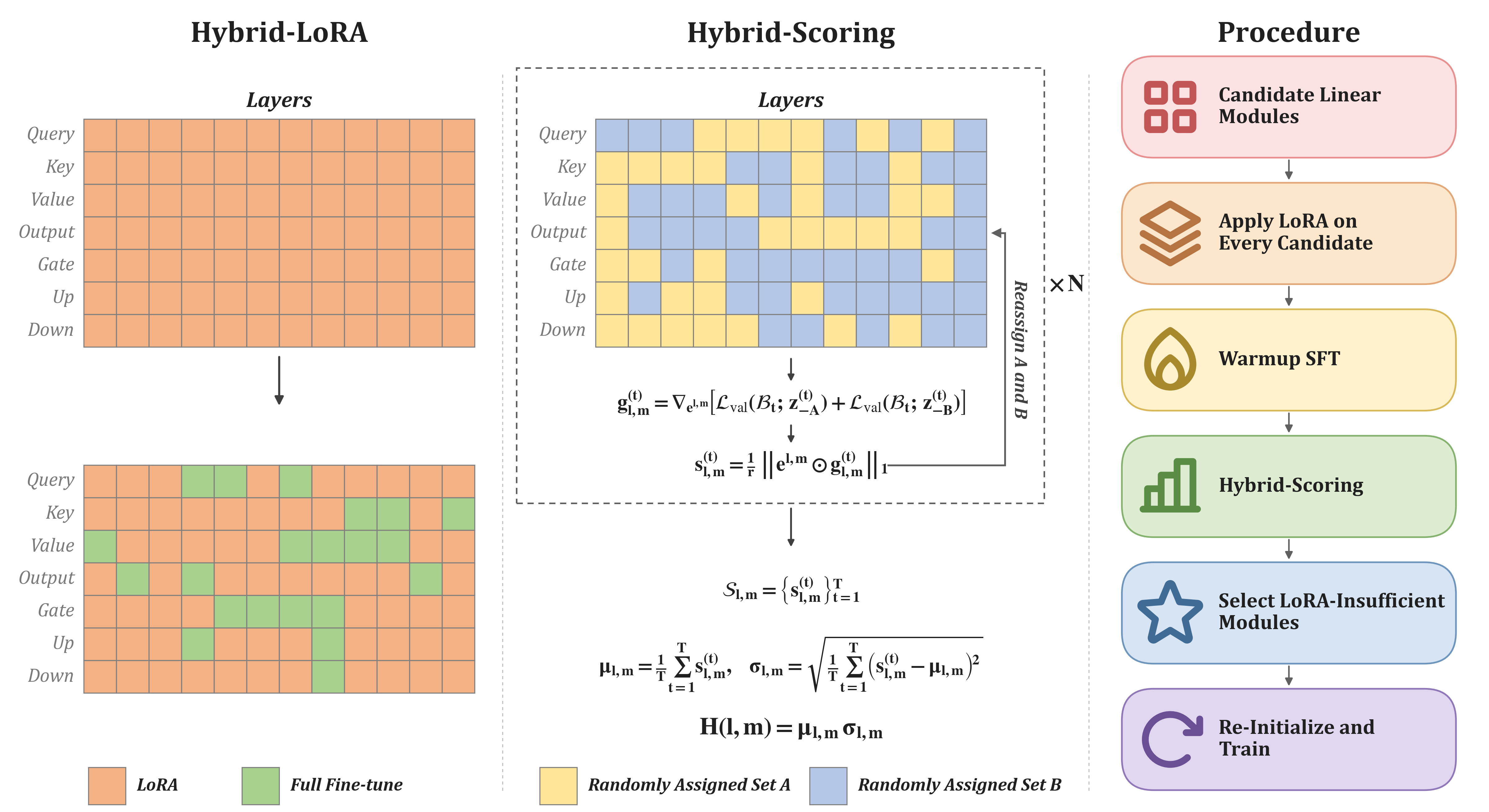}
    \caption{Overview of the Hybrid-LoRA framework. A LoRA-based probing stage is used to score candidate linear modules, after which selected modules are trained with full fine-tuning and the remaining modules are adapted with LoRA.}
    \label{fig:overview}
\end{figure}

We present Hybrid-LoRA, a two-stage framework for efficient LLM post-training that first scores candidate Transformer linear modules through a short LoRA-based probing stage and then allocates a small full fine-tuning budget while adapting the remaining modules with LoRA. As shown in Figure~\ref{fig:overview}, Hybrid-LoRA consists of two stages. In the probing stage, we attach LoRA modules to all candidate linear modules and run a short warm-up training phase. We then compute Hybrid-Score to rank modules by their gradient-weighted sensitivity. In the final training stage, we reinitialize the model, assign low-score modules to full fine-tuning under the parameter budget, and adapt the remaining modules with LoRA. We provide preliminaries on RLVR, GRPO, and GSPO in Appendix~\ref{app: preliminaries}.

\subsection{Formulation: Hybrid Adaptation under a Trainable Parameter Budget}
\label{sec: formulation}
Given a Transformer-based LLM~\citep{vaswani2017attention} with $L$ layers, we consider the following set of candidate linear module types eligible for adaptation:
\begin{equation}
\label{eq:candidate_modules}
\mathcal{A}=\{W_q,W_k,W_v,W_o,W_{\mathrm{gate}},W_{\mathrm{up}},W_{\mathrm{down}}\}.
\end{equation}
We define the set of all candidate linear modules as
\begin{equation}
\label{eq:candidate_module_set}
\mathcal{U} = \{1,\dots,L\} \times \mathcal{A}.
\end{equation}
Here, $(l,m)\in\mathcal{U}$ denotes the module of type $m$ in layer $l$.

We aim to partition $\mathcal{U}$ into two disjoint subsets,
\begin{equation}
\label{eq:module_partition}
\mathcal{S}_{\mathrm{fft}} \cup \mathcal{S}_{\mathrm{lora}} = \mathcal{U}, 
\qquad
\mathcal{S}_{\mathrm{fft}} \cap \mathcal{S}_{\mathrm{lora}} = \emptyset.
\end{equation}
where $\mathcal{S}_{\mathrm{fft}}$ and $\mathcal{S}_{\mathrm{lora}}$ denote the modules assigned to FFT and LoRA adaptation, respectively. Let $P_{l,m}$ denote the number of trainable parameters in module $(l,m)$ under FFT. Given an FFT budget ratio $R_{\mathrm{fft}}\in(0,1)$, we impose the constraint
\begin{equation}
\label{eq:fft_budget_constraint}
\frac{\sum_{(l,m)\in \mathcal{S}_{\mathrm{fft}}} P_{l,m}}
{\sum_{(l,m)\in \mathcal{U}} P_{l,m}}
\le R_{\mathrm{fft}}.
\end{equation}
Our objective is to find a partition $(\mathcal{S}_{\mathrm{fft}}, \mathcal{S}_{\mathrm{lora}})$ that maximizes the post-training performance of the model $\mathcal{M}$:
\begin{equation}
\label{eq:hybrid_allocation_objective}
\max_{\mathcal{S}_{\mathrm{fft}}, \mathcal{S}_{\mathrm{lora}}}
\;\; \mathcal{J}\big(\mathcal{M}; \mathcal{S}_{\mathrm{fft}}, \mathcal{S}_{\mathrm{lora}}\big)
\quad
\text{s.t. Equation~\ref{eq:fft_budget_constraint}.}
\end{equation}

This formulation defines a combinatorial optimization problem over module assignments, which is intractable to solve exactly due to the exponential size of the search space. Therefore, an efficient approximation strategy is required in practice.

\subsection{LoRA Re-parameterization and Architecture Parameters}
\label{sec:lora_reparam}

As defined in Equation~\ref{eq:candidate_modules}, we consider seven candidate linear module types in each Transformer layer. During the probing stage, we apply LoRA to all candidate modules in each layer. For a given input activation $x$ to module $m$ in layer $l$, the LoRA-adapted output is written as
\begin{equation}
\label{eq:lora_adapted_output}
x' = xW^{l,m} + b^{l,m} + \alpha^{l,m} \cdot x A^{l,m} B^{l,m}.
\end{equation}
where $W^{l,m}$ and $b^{l,m}$ denote the frozen pretrained weight and bias of the original linear module, respectively. $A^{l,m} \in \mathbb{R}^{d_{\mathrm{model}} \times r}$ and $B^{l,m} \in \mathbb{R}^{r \times d_{\mathrm{out}}}$ are the LoRA low-rank matrices, and $\alpha^{l,m}$ is a module-specific architecture parameter that controls the contribution of the LoRA branch. To facilitate the Hybrid-Score calculation, we follow the SVDLinear parameterization used in AdaLoRA~\citep{zhang2023adaptive} and decompose the LoRA update into a singular-value-style form:
\begin{equation}
\label{eq:svd_lora_reparam}
x' = xW^{l,m} + b^{l,m} + \alpha^{l,m} \cdot x A^{l,m} E^{l,m} B^{l,m},
\end{equation}
where $E^{l,m}=\operatorname{diag}(\boldsymbol{e}^{l,m})$ is the diagonal coefficient
matrix in the SVD-style LoRA parameterization, and
$\boldsymbol{e}^{l,m}\in\mathbb{R}^{r}$ contains the singular-value-style coefficients over the $r$ LoRA components.

\subsection{Hybrid-Score Calculation}
\label{sec:hybrid_score}
\paragraph{Limitations of existing importance estimators.}
A straightforward way to estimate module importance is to use the magnitude of the learned architecture parameter as a proxy:
\begin{equation}
\label{eq:alpha_importance}
I_{\alpha}(l,m)=|\alpha^{l,m}|.
\end{equation}
This follows the spirit of differentiable architecture search~\citep{liu2018darts} and adaptive LoRA allocation~\citep{zhang2024autolora}. However, learned architecture-parameter magnitudes may not faithfully reflect the true contribution of an operation~\citep{xiao2022shapley}, making \(|\alpha^{l,m}|\) alone insufficient.

A more faithful estimator is perturbation-based scoring, which measures the validation-loss change after removing a module~\citep{ghorbani2019data,xiao2022shapley}:
\begin{equation}
\label{eq:module_perturbation_score}
P(l,m)=
\mathcal{L}_{\mathrm{val}}(\theta_{\setminus(l,m)})
-
\mathcal{L}_{\mathrm{val}}(\theta).
\end{equation}
However, computing this score for all candidate modules requires $
1+|\mathcal{U}|=1+|\mathcal{A}|\cdot L$
validation forward passes per batch, which is prohibitively expensive for LLMs. Therefore, we introduce an attribution-based Hybrid-Score to approximate module contribution efficiently.

\paragraph{Hybrid-Score.} We propose Hybrid-Score to estimate how effectively each candidate linear module can be adapted by its LoRA branch after the LoRA-based probing stage. The overall procedure is summarized in  Appendix~\ref{app:algorithms}. Let the validation set be divided into $T$ mini-batches $\{\mathcal{B}_t\}_{t=1}^{T}$. For each validation batch $\mathcal{B}_t$, we randomly partition the candidate module set into two complementary buckets,
\begin{equation}
\label{eq:random_bucket_partition}
\mathcal{U}=\mathcal{U}_{A}^{(t)} \cup \mathcal{U}_{B}^{(t)},
\qquad
\mathcal{U}_{A}^{(t)} \cap \mathcal{U}_{B}^{(t)}=\emptyset.
\end{equation}
We then perform two masked backward passes: one with $\mathcal{U}_{A}^{(t)}$ disabled and the other with $\mathcal{U}_{B}^{(t)}$ disabled. Let $z_{-A}^{(t)}$ denote the mask that disables modules in $\mathcal{U}_{A}^{(t)}$, and $z_{-B}^{(t)}$ denote the mask that disables modules in $\mathcal{U}_{B}^{(t)}$. For module $(l,m)$, we compute the gradient with respect to the SVD-style diagonal
coefficient vector $\boldsymbol{e}^{l,m}$ introduced in Equation~\ref{eq:svd_lora_reparam}:
\begin{equation}
\label{eq:hybrid_score_gradient}
\boldsymbol{g}_{l,m}^{(t)}
=
\nabla_{\boldsymbol{e}^{l,m}}
\left[
\mathcal{L}_{\mathrm{val}}(\mathcal{B}_t; z_{-A}^{(t)})
+
\mathcal{L}_{\mathrm{val}}(\mathcal{B}_t; z_{-B}^{(t)})
\right].
\end{equation}
Since module $(l,m)$ is disabled in exactly one of the two passes, only the pass where it remains active contributes a non-zero gradient to $g_{l,m}^{(t)}$.

The batch-level sensitivity score is then computed as
\begin{equation}
\label{eq:batch_sensitivity_score}
s_{l,m}^{(t)}
=
\frac{1}{r}
\left\|
\boldsymbol{e}^{l,m}
\odot
\boldsymbol{g}_{l,m}^{(t)}
\right\|_1 .
\end{equation}
Here, $\odot$ denotes element-wise multiplication, so
$\boldsymbol{e}^{l,m}\odot\boldsymbol{g}_{l,m}^{(t)}$
measures the dimension-wise first-order contribution of the LoRA coefficients. Across all validation batches, we collect a sensitivity list for each module:
\begin{equation}
\mathcal{S}_{l,m}
=
\left\{
s_{l,m}^{(t)}
\right\}_{t=1}^{T}.
\end{equation}
The final Hybrid-Score is obtained by aggregating this list:
\begin{equation}\label{eq:aggregated_score}
H(l,m)=\mu_{l,m}\cdot\sigma_{l,m},
\end{equation}
where
\begin{equation}
\label{eq:score_mean_std}
\mu_{l,m}
=
\frac{1}{T}
\sum_{t=1}^{T}
s_{l,m}^{(t)},
\qquad
\sigma_{l,m}
=
\sqrt{
\frac{1}{T}
\sum_{t=1}^{T}
\left(
s_{l,m}^{(t)}-\mu_{l,m}
\right)^2
}.
\end{equation}

The random bucket masking is inspired by Shapley-NAS and R-Drop~\citep{xiao2022shapley,wu2021r}, where importance is estimated under sampled contexts rather than a single fixed architecture. This design accounts for interactions among LoRA modules, since the contribution of one module may depend on which other modules are active. For each validation batch, two complementary masked backward passes ensure that every module is active in one pass and disabled in the other, yielding one batch-level sensitivity value for each module under a sampled co-active context.

Following first-order sensitivity criteria for network pruning~\citep{lee2018snip}, we define Hybrid-Score as a module-level diagnostic of how effectively each module is adapted by its LoRA branch. Specifically, $\boldsymbol{e}^{l,m}\odot \boldsymbol{g}_{l,m}^{(t)}$ approximates the first-order effect of perturbing the SVD-style diagonal coefficients of module $(l,m)$ on the validation loss. The mean term $\mu_{l,m}$ measures average LoRA-side sensitivity across validation batches, while the standard deviation term $\sigma_{l,m}$ captures variation across sampled co-active contexts, reflecting context-dependent interactions.

Therefore, a high Hybrid-Score indicates that the LoRA branch of module $(l,m)$ has a strong and consistent influence on validation behavior under diverse masking contexts, suggesting that LoRA is sufficient for this module. Conversely, a low Hybrid-Score suggests weak LoRA-side adaptation sensitivity, implying that low-rank adaptation may be insufficient. We thus assign low-score modules to full fine-tuning and keep high-score modules under LoRA adaptation.

\subsection{Training Algorithm}
\label{sec:training_algorithm}

The overall training procedure of Hybrid-LoRA is provided in Appendix~\ref{app:algorithms}. The training process consists of two stages. In the probing stage, we attach LoRA modules to all candidate linear modules in $\mathcal{U}$ from Equation~\ref{eq:candidate_module_set} and freeze the pretrained weights. After a short warmup period, the probing model is used to estimate the Hybrid-Score $H(l,m)$ in Equation~\ref{eq:aggregated_score} for each candidate module on the validation set.

The resulting scores are then used for budget-constrained hybrid module allocation. Specifically, we sort all candidate modules in ascending order of $H(l,m)$ and greedily add the lowest-score modules into $\mathcal{S}_{\mathrm{fft}}$ until the budget constraint in Equation~\ref{eq:fft_budget_constraint} is reached. The remaining modules are assigned to $\mathcal{S}_{\mathrm{lora}}$ according to the disjoint partition in Equation~\ref{eq:module_partition}. This provides an efficient approximation to the hybrid allocation objective in Equation~\ref{eq:hybrid_allocation_objective}.

In the formal training stage, we reinitialize the model from the original pretrained checkpoint $\mathcal{M}_0$ and apply the selected hybrid allocation. Modules in $\mathcal{S}_{\mathrm{fft}}$ are fully fine-tuned, while modules in $\mathcal{S}_{\mathrm{lora}}$ are adapted with LoRA. The final model is then optimized with the standard post-training objective under this mixed parameterization. This design separates module selection from final optimization: the probing stage only determines the adaptation allocation, while the final performance is obtained from a freshly initialized hybrid model.

\section{Experiments}
\label{sec: experiments}
In this section, we conduct a series of experiments to evaluate our Hybrid-LoRA method.

\subsection{Models}
We compare our Hybrid-LoRA framework with the current state-of-the-art LoRA baselines on Qwen-2.5 1.5B, 3B and 7B Instruct.  In particular, we compare against IO and CoT prompting baselines, full fine-tuning (FFT), and representative PEFT methods including LoRA~\citep{hu2022lora}, AdaLoRA~\citep{zhang2023adaptive}, AutoLoRA~\citep{zhang2024autolora}, and LoRA-Drop~\citep{zhou2025lora}. The baselines are implemented based on their open-sourced codes.

\subsection{Datasets and Evaluation Protocols}

\paragraph{Datasets.}
We evaluate Hybrid-LoRA on six public reasoning benchmarks covering mathematics, knowledge-intensive question answering, and code generation. Specifically, we use Math-500~\citep{lightman2023let}, AIME-24 and AIME-25~\citep{maa_aime}, MMLU-Pro~\citep{wang2024mmlu}, GPQA~\citep{rein2023gpqa}, and LeetCodeDataset~\citep{xia2025leetcodedataset}. Math-500 and AIME evaluate multi-step mathematical reasoning and competition-level problem solving; MMLU-Pro and GPQA evaluate challenging academic and professional knowledge reasoning; and LeetCodeDataset measures program synthesis ability.

For post-training, we use Mixture-of-Thoughts~\citep{openr1} as the unified training corpus. It contains approximately 350K verified reasoning trajectories distilled from DeepSeek-R1, providing diverse and structured reasoning supervision for RLVR/GRPO-style post-training. Using a single post-training corpus allows us to compare different adaptation strategies under the same data setting.

\paragraph{Evaluation Protocols.}
We use task-appropriate metrics for each benchmark. For mathematical reasoning, we report \texttt{pass@1} on Math-500 and \texttt{pass@4} on AIME-24 and AIME-25, following prior reasoning-oriented evaluation protocols~\citep{yang2025qwen3}. Under \texttt{pass@k}, each problem is sampled $k$ times and is considered solved if any generated solution matches the reference answer.

For LeetCodeDataset, we report \texttt{pass@4}, where each problem is evaluated with four independently generated programs and is counted as correct if any program passes the corresponding test cases. For MMLU-Pro and GPQA, we use a single-response setting and report accuracy based on the standard answer-matching rule of each benchmark.

\subsection{Training Details}
All experiments are conducted on 8 NVIDIA A800 80GB GPUs. We use the same training hyperparameters for all post-training baselines and our Hybrid-LoRA pipeline unless otherwise specified. The sampling temperature is set to $T=1.0$. We use a learning rate of $1\times10^{-6}$ for full fine-tuned modules and $2.5\times10^{-5}$ for LoRA modules. For the reinforcement learning stage, we use a group size of 4, a clipping range of $\pm 0.2$, and a KL penalty coefficient of 0.01. For both Hybrid-Score estimation and Hybrid-LoRA training, LoRA modules are initialized with rank $r=16$. The probing warm-up stage is trained for 500 steps. We set the full fine-tuning module ratio to $R_{\mathrm{fft}}=10\%$ and use $K=20$ random partitions for Hybrid-Score estimation. During the final training stage, we evaluate the model every 50 steps and keep the checkpoint with the lowest validation loss.

\subsection{Baselines}
We compare Hybrid-LoRA with both inference-only and post-training baselines. In total, we consider seven baselines, including two inference-only baselines, one FFT reference, and four PEFT or hybrid fine-tuning baselines. For inference-only baselines, \textbf{Input--Output prompting (IO)} directly prompts the LLM to generate the final answer without additional training or explicit reasoning instructions, while \textbf{Chain-of-Thought prompting (CoT)} asks the model to produce intermediate reasoning steps before giving the final answer, also without any parameter updates.

For post-training baselines, \textbf{Full-FT} updates all parameters of the LLM and serves as the upper-bound reference under full fine-tuning. We also compare with several PEFT-based module selection baselines. Specifically, we adapt the importance criteria used in AdaLoRA, AutoLoRA, and LoRA-Drop to identify linear modules whose LoRA adaptation appears insufficient under each criterion. For \textbf{AdaLoRA}, modules with the smallest allocated ranks after adaptive rank allocation are considered to receive limited LoRA capacity. For \textbf{AdaLoRA}, \textbf{AutoLoRA}, and \textbf{LoRA-Drop}, we rank modules according to their corresponding LoRA-side criteria: allocated rank, learned architecture weight $\alpha$, and LoRA output contribution, respectively. Modules with the \textbf{lowest scores} under each criterion are treated as less effectively adapted by LoRA and are assigned to full fine-tuning. In each case, we select the bottom $10\%$ of linear modules according to the corresponding LoRA-side criterion and assign these modules to FFT, while applying LoRA to the remaining modules. This design keeps the number of tunable parameters comparable to our Hybrid-LoRA setting and allows a controlled comparison of different module-selection strategies.

Finally, we include \textbf{LoRA-FFN}, a heuristic hybrid baseline that applies LoRA to all FFN linear layers while full fine-tuning all attention linear modules.

\subsection{Main Results}
The experimental results on the six reasoning benchmarks are presented in Table \ref{tab:main_results}. In the first three columns, we present the post-training approach, parameter efficient approach and number of tunable parameters respectively. Input/output prompting (IO) and Chain-of-Thought (CoT) are included as inference-only baselines without any post-training for reference.

Overall, our proposed \textbf{Hybrid-LoRA} consistently outperforms the benchmark-wise best baseline by an average of 4.36\% across all model scales and benchmarks, while using only a small fraction of the trainable parameters required by FFT. On Qwen-2.5 1.5B, Hybrid-LoRA attains performance comparable to or better than FFT under both GRPO and GSPO settings, while reducing the number of tunable parameters from 1544M to around 105M. Similar trends are observed on the 3B and 7B models, where Hybrid-LoRA matches or slightly outperforms FFT on most benchmarks, particularly on challenging reasoning tasks such as AIME and GPQA.

\begin{table*}[ht]
\newcommand{\thinsep}{\specialrule{0.3pt}{1pt}{1pt}}
\centering
\caption{Overall reasoning benchmark performance. Full-FT (ref.) is the full-parameter reference.}
\label{tab:main_results}
\resizebox{0.99\textwidth}{!}{
\renewcommand\arraystretch{1.05}
\begin{tabular}{ll|c|cccccc}
\hline
\textbf{Post-training}   &    \textbf{Param efficient}    &   \textbf{\#Tunable}    & Math-500  &  AIME-24   &  AIME-25    &  MMLU-Pro   &  GPQA   &     LeetCodeDataset       \\ 
\textbf{Approach}      &   \textbf{Approach}     &    \textbf{Params}     &    pass@1   &  pass@4   &  pass@4   &    Accuracy       &   Accuracy    &  pass@4     \\

\hline
\multicolumn{9}{l}{\textbf{\emph{Qwen-2.5 1.5B Instruct}}}  \\
\hline
IO    &   -   &   0.0   &   25.4  &   3.3     &  0.0   &   13.1    &   18.8       &  0.0 \\
COT   &   -   &   0.0   &     56.8   &    14.7   &   10.7     &   20.6    &     34.1   &    14.6 \\  
SFT    &    Full-FT   &    1544M    &    68.4   &    22.7    &  21.3    &     46.9   &   37.9       &   40.6                  \\  
GRPO  &   Full-FT (ref.)   &   1544M   &     80.6    &    40.0    &   36.7     &  55.6    &    37.2    &     48.3       \\
GRPO  &    AdaLoRA    &    105M    &    75.3     &    32.0   &    33.3    &       50.5    &  33.9     &  45.4   \\
GRPO  &    AutoLoRA    &    106M     &   77.4     &   34.7   &   31.3       &  52.1   &  34.4    &  45.7   \\
GRPO  &    LoRA-Drop    &    105M    &   76.8     &    34.0    &   32.0    &  53.2   &  35.1    &   46.3    \\
GRPO  &    Hybrid-LoRA (ours)    &    105M    &    \textbf{80.2}     &   \textbf{38.0}    &    \textbf{35.3}    &  \textbf{54.9}    &  \textbf{37.3}      &     \textbf{48.0}   \\
\thinsep
GSPO   &   Full-FT (ref.)   &   1544M   &      82.0   &      42.0    &   37.3    &   56.8   &   38.7  &       49.1     \\
GSPO   &    LoRA-FFN     &    122M    &     79.2     &   34.0    &     30.7    &      53.1     &    36.3    &  44.5     \\
GSPO  &    AdaLoRA    &    106M    &     79.4     &   36.0    &   31.3     &   53.5    &  36.6    &   44.8    \\
GSPO  &    AutoLoRA    &    107M    &    80.2   &   35.3   &     32.7     &    54.9    &   37.1   &  45.2  \\
GSPO  &    LoRA-Drop    &    105M     &    79.8    &    40.0    &     34.0    &      54.7     &   36.9    &   44.7   \\
GSPO  &     Hybrid-LoRA (ours)          &    105M      &    \textbf{81.6}    &   \textbf{42.0}    &   \textbf{36.7}    &      \textbf{56.9}   &     \textbf{38.5}   &      \textbf{47.6}     \\

\hline
\multicolumn{9}{l}{\textbf{\emph{Qwen-2.5 3B Instruct}}}  \\
\hline
CoT   &    -   &     0.0   &  77.2   &     24.0    &   20.6      &     40.4      &     50.3    &       40.2    \\  
GRPO  &   Full-FT (ref.)   &   3086M   &    91.2   &     46.0    &   44.0    &   71.8     &   60.7    &       63.8     \\
GRPO  &    AutoLoRA    &    201M   &     89.6    &     40.7   &  38.7   &    69.3     &     58.4    &    59.6       \\
GRPO  &    Hybrid-LoRA (ours)    &     192M   &    \textbf{91.4}    &    \textbf{43.3}    &    \textbf{42.0}    &    \textbf{71.9} &    \textbf{60.4}    &    \textbf{62.7}   \\
\thinsep
GSPO   &   Full-FT (ref.)   &   3086M   &  92.4  &   49.3      &   47.3   &     73.4    &   62.1   &      65.1   \\
GSPO  &    AutoLoRA    &   194M   &    90.2   &    44.0    &    40.7   &   71.1    &    60.3    &  62.6        \\
GSPO  &    Hybrid-LoRA (ours)    &     192M   &    \textbf{91.8}    &   \textbf{48.0}   &  \textbf{45.3}    &     \textbf{72.9}   &    \textbf{61.6}    &    \textbf{64.0}  \\

\hline
\multicolumn{9}{l}{\textbf{\emph{Qwen-2.5 7B Instruct}}}  \\
\hline

GRPO  &    AutoLoRA    &    453M   &    91.4   &     52.7    &    48.7     &    80.3    &    73.5     &     71.5        \\
GRPO  &    Hybrid-LoRA (ours)    &  449M   &       \textbf{93.6}  &   \textbf{56.0}   &   \textbf{52.0}    &   \textbf{83.1}    &   \textbf{76.3}    &   \textbf{73.2}    \\
\thinsep

GSPO  &    AutoLoRA    &    453M   &    92.2    &    55.3   &   52.7    &  81.9    &    74.8    &  72.5     \\
GSPO  &    Hybrid-LoRA (ours)    &   449M   &     \textbf{94.4}     &   \textbf{60.0}   &  \textbf{55.3}    &    \textbf{83.8}    &   \textbf{77.9}    &      \textbf{74.5}     \\
\hline
\end{tabular}}
\end{table*}

Compared with existing parameter-efficient methods, including AdaLoRA, AutoLoRA, and LoRA-Drop, Hybrid-LoRA consistently delivers superior or more stable performance. This demonstrates the advantage of selectively allocating full fine-tuning capacity to modules that are less effectively adapted by LoRA, instead of uniformly applying low-rank adaptation. Notably, Hybrid-LoRA shows clear improvements on math-intensive benchmarks (e.g., Math-500, AIME-24/25), suggesting that accurate module importance estimation is particularly beneficial for complex reasoning tasks.

Furthermore, the results indicate that simply increasing the number of trainable parameters in PEFT methods does not necessarily close the performance gap with FFT. Instead, \textbf{how the parameters are allocated across model modules plays a more critical role}. Hybrid-LoRA effectively bridges this gap by combining the expressiveness of FFT with the efficiency of LoRA, achieving near-FFT performance with significantly reduced computational cost.

\begin{figure*}[t]
\centering
\includegraphics[width=0.86\linewidth]{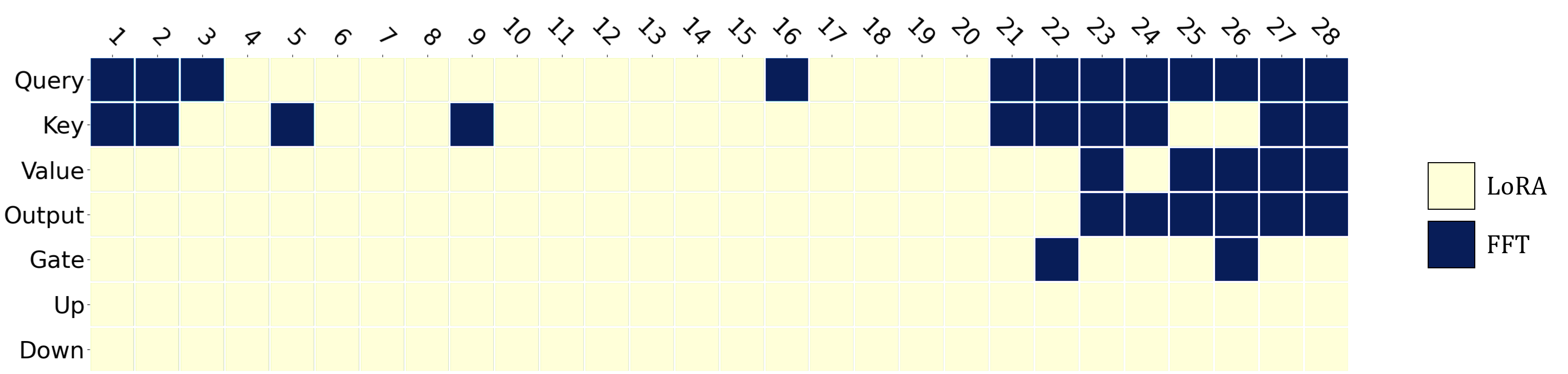}
\caption{The LoRA module placements on the Qwen2.5 1.5B backbones, after post-trained on the Mixture-of-thoughts. The linear modules without LoRA modifications (shown as the red cells) are fine-tuned in full parameters. }
\label{fig:015b_module_allocation}
\vspace{-8pt}
\end{figure*}

\subsection{Module Distribution Analysis}
Figure~\ref{fig:015b_module_allocation} visualizes the final module allocation obtained by our Hybrid-Score-based pipeline. The selected full fine-tuned modules with low Hybrid-Scores are concentrated in the attention blocks of the later layers, especially the query, key, value, and output projections. In contrast, most FFN modules, including the gate, up, and down projections, remain assigned to LoRA adaptation. This allocation pattern suggests that, under the same parameter budget, the later attention projections are more sensitive to the choice of adaptation strategy. A possible explanation is that these modules are
directly involved in refining high-level token interactions and reasoning patterns during post-training, where the limited low-rank update space of LoRA may be less expressive than full-parameter updates. Meanwhile, FFN modules appear to be more effectively handled by LoRA, indicating that low-rank adaptation is sufficient for most feed-forward transformations in this setting.

\subsection{Ablation Studies}
\begin{figure}[t]
\centering
\begin{subfigure}[b]{0.48\textwidth}
\centering
\includegraphics[width=\linewidth]{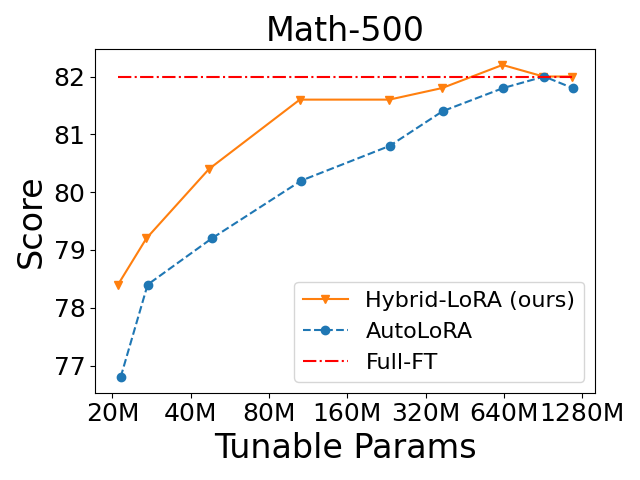}
\caption{Performance on Math-500}
\label{subfig:Math-500_different_tunable_paras}
\end{subfigure}
\vspace{-5pt}
\begin{subfigure}[b]{0.48\textwidth}
\centering
\includegraphics[width=\linewidth]{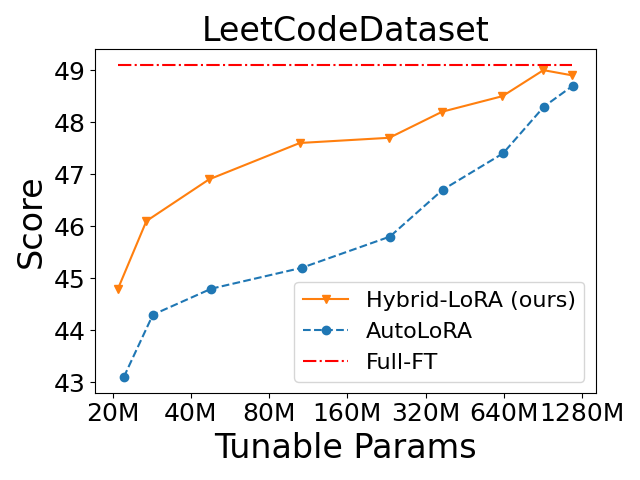}
\caption{Performance on LeetCodeDataset }
\label{subfig:LeetCodeDataset_different_tunable_paras}
\end{subfigure}
\caption{Ablation studies on the number of tunable parameters. }
\label{fig:params_to_scores}
\vspace{-8pt}
\end{figure}

\paragraph{Tunable Parameters.} We ablate the effect of tunable parameter budgets in Figure~\ref{fig:params_to_scores}, using AutoLoRA as the comparison baseline. As the number of tunable parameters increases, Hybrid-LoRA consistently improves faster than AutoLoRA on both Math-500 and LeetCodeDataset, indicating that our Hybrid-Score selects modules that benefit more effectively from full fine-tuning.

\paragraph{Hybrid-Score Variants.} We further evaluate several variants of the Hybrid-Score aggregation rule in Table~\ref{tab:hybrid_score_ablation}. 
Hybrid-LoRA-1 replaces the product form with a mean-to-variability ratio $\mu_{l,m}/\sigma_{l,m}$, while Hybrid-LoRA-2 uses the coefficient-of-variation-style form $\sigma_{l,m}/\mu_{l,m}$. 
Hybrid-LoRA-3 adopts a mean-variance-style additive score $\mu_{l,m}+0.5\sigma_{l,m}$.
The original Hybrid-LoRA and Hybrid-LoRA-4 achieve comparable performance across the two benchmarks. However, Hybrid-LoRA-4 is much less practical because it starts from a full fine-tuning configuration and replaces lower-scoring modules with LoRA, leading to substantially higher allocation cost, running time, and memory usage. In contrast, the original Hybrid-LoRA obtains similar performance through an efficient LoRA-probing stage, making it the more practical design.

\begin{table*}[ht]
\centering
\caption{Ablation study of different Hybrid-Score variants on reasoning benchmarks.}
\label{tab:hybrid_score_ablation}
\resizebox{0.65\textwidth}{!}{
\renewcommand\arraystretch{1.05}
\begin{tabular}{l|cc}
\hline
\textbf{Method} & Math-500 & LeetCodeDataset \\
                & pass@1   & pass@4          \\
\hline
Hybrid-LoRA 
& \textbf{81.6} & 47.6 \\

Hybrid-LoRA-1 ($\mu_{l,m}/\sigma_{l,m}$) 
& 80.8 & 46.9 \\

Hybrid-LoRA-2 ($\sigma_{l,m}/\mu_{l,m}$) 
& 79.4 & 45.3 \\

Hybrid-LoRA-3 ($\mu_{l,m} + 0.5 \times \sigma_{l,m}$) 
& 81.2 & 46.4 \\

Hybrid-LoRA-4 (pruning from Full-FT) 
& 81.4 & \textbf{47.7} \\

\hline
\end{tabular}}
\end{table*}

\section{Conclusion and Limitations}
\label{sec:conclusion}
In this paper, we propose Hybrid-LoRA, a heterogeneous adaptation framework that assigns modules less effectively adapted by LoRA to full fine-tuning while adapting the remaining modules with LoRA. We introduce Hybrid-Score, a module-importance metric computed from tunable parameter values and gradients, which provides a reliable allocation criterion beyond prior bi-level optimization or heuristic scoring methods. Across multiple complex reasoning tasks and LLMs of different scales, Hybrid-LoRA closely matches full fine-tuning performance under a 10\% full fine-tuning module budget, while adapting the remaining candidate modules with LoRA.

Our work has limitations. Due to computational constraints, we scale experiments only up to 7B models. 
In addition, our current method follows a two-stage probing-and-training pipeline; future work may explore end-to-end or dynamically updated allocation strategies.

\bibliographystyle{plainnat}
\bibliography{references}

\newpage
\appendix
\section{Broader Impact}
Hybrid-LoRA aims to reduce the computational and memory cost of reasoning-oriented LLM post-training by combining full fine-tuning and LoRA adaptation under a fixed parameter budget. This can make advanced post-training more accessible to researchers and organizations with limited computational resources, and may reduce the cost of adapting LLMs for scientific, educational, and engineering applications. However, more efficient fine-tuning methods may also lower the barrier for adapting LLMs to harmful or misleading uses, such as generating misinformation, spam, or unsafe instruction-following behaviors. In addition, although Hybrid-LoRA improves parameter efficiency, large-scale LLM training and evaluation still incur non-negligible computational and environmental costs. These considerations highlight the need for responsible dataset selection, careful evaluation, and appropriate deployment safeguards when applying efficient post-training methods.
\section{Preliminaries}
\label{app: preliminaries}
\subsection{Reinforcement Learning with Verifiable Rewards (RLVR) for Reasoning Post-Training}
We consider reinforcement learning with verifiable rewards (RLVR) for reasoning-oriented post-training of large language models. In this setting, a pretrained language model, typically initialized from an SFT checkpoint, receives an input problem $x$ sampled from a dataset $\mathcal{D}$ and generates an output sequence $y \sim \pi_\theta(\cdot \mid x)$. The output may include both an intermediate reasoning process and a final answer. Unlike preference-based reinforcement learning, RLVR uses a rule-based or programmatic verifier that assigns a reward according to whether the generated answer is correct or otherwise verifiable.

Formally, for each input-output pair $(x, y)$, a verifier defines a reward function
\begin{equation}
\label{eq:reward_function}
r(x,y)\in\mathbb{R}.
\end{equation}
which is typically sparse and depends on the correctness of the final answer, although more structured rewards can also be used when parts of the reasoning process are verifiable. The RLVR objective is to optimize the policy $\pi_\theta$ so as to maximize the expected reward:
\begin{equation}
\label{eq:rlvr_objective}
\max_\theta \; \mathbb{E}_{x \sim \mathcal{D},\, y \sim \pi_\theta(\cdot \mid x)} [r(x, y)].
\end{equation}
\subsection{GRPO for RLVR Optimization}
\label{sec:grpo}
Group Relative Policy Optimization (GRPO) is a critic-free policy optimization
method widely used for RLVR-based reasoning post-training. Given a prompt $x$,
GRPO samples a group of responses $\{y_i\}_{i=1}^{G}$ from the current policy
and assigns each response a scalar reward $r_i$. Instead of training a separate
value model, GRPO estimates the advantage of each response by normalizing its
reward within the sampled group:
\begin{equation}
\label{eq:group_normalized_advantage}
A_i=\frac{r_i-\mu_r}{\sigma_r},
\qquad
\mu_r=\frac{1}{G}\sum_{i=1}^G r_i,
\qquad
\sigma_r=\sqrt{\frac{1}{G}\sum_{i=1}^G (r_i-\mu_r)^2+\epsilon}.
\end{equation}
The policy is then updated with a PPO-style clipped objective and a KL penalty
against a reference policy. In RLVR, the reward is typically constructed from
verifiable signals such as answer correctness, format compliance, and length
control. 
Given a prompt $x$, GRPO samples a group of $G$ responses
$\{y_i\}_{i=1}^{G}$ from the behavior policy
$\pi_{\theta_{\mathrm{old}}}$. Each response is assigned a scalar reward $r_i$,
and the group-relative advantage is computed as
\[
A_i=\frac{r_i-\mu_r}{\sigma_r},
\qquad
\mu_r=\frac{1}{G}\sum_{i=1}^G r_i,
\qquad
\sigma_r=\sqrt{\frac{1}{G}\sum_{i=1}^G (r_i-\mu_r)^2+\epsilon}.
\]

For each token $y_{i,t}$, the importance ratio between the updated policy
$\pi_\theta$ and the behavior policy $\pi_{\theta_{\mathrm{old}}}$ is
\[
\rho_{i,t}(\theta)
=
\frac{\pi_\theta(y_{i,t}\mid x,y_{i,<t})}
{\pi_{\theta_{\mathrm{old}}}(y_{i,t}\mid x,y_{i,<t})}.
\]
GRPO optimizes a PPO-style clipped surrogate objective:
\[
\ell_{i,t}(\theta)
=
\min\!\Big(
\rho_{i,t}(\theta)A_i,\;
\mathrm{clip}\!\big(\rho_{i,t}(\theta),1-\epsilon_{\mathrm{clip}},1+\epsilon_{\mathrm{clip}}\big)A_i
\Big).
\]
The final GRPO objective is
\[
\mathcal{J}_{\mathrm{GRPO}}(\theta)
=
\mathbb{E}_{x,\{y_i\}}
\left[
\frac{1}{G}\sum_{i=1}^G
\frac{1}{|y_i|}
\sum_{t=1}^{|y_i|}
\ell_{i,t}(\theta)
\right]
-
\beta\,
\mathrm{KL}\!\left(
\pi_\theta \,\|\, \pi_{\mathrm{ref}}
\right),
\]
where $\pi_{\mathrm{ref}}$ denotes the reference policy and $\beta$ controls the
strength of the KL regularization.

\section{Algorithms}
\label{app:algorithms}
We provide the detailed procedures of Hybrid-LoRA in Algorithm~\ref{alg:hybrid_score} and Algorithm~\ref{alg:hybrid_lora_overview}. 
Algorithm~\ref{alg:hybrid_score} presents the Hybrid-Score calculation, which aggregates gradient-weighted LoRA-side sensitivities over validation batches and randomly sampled masking contexts. 
Algorithm~\ref{alg:hybrid_lora_overview} gives the full Hybrid-LoRA training pipeline: the model is first warmed up with LoRA on all candidate modules, then modules are ranked by their Hybrid-Scores, and finally low-scoring modules are assigned to full fine-tuning while the remaining modules are trained with LoRA.
\begin{algorithm}
\caption{Hybrid-Score Calculation}
\label{alg:hybrid_score}
\begin{algorithmic}[1]
\Require Probing model $\mathcal{M}_{\mathrm{probe}}$, candidate module set $\mathcal{U}$, validation dataloader $\mathcal{D}_{\mathrm{val}}$
\Ensure Hybrid-Score $H(l,m)$ for each $(l,m)\in\mathcal{U}$

\State Initialize $\mathcal{S}_{l,m}\leftarrow [\ ]$ for each $(l,m)\in\mathcal{U}$

\For{each validation batch $\mathcal{B}_t\in\mathcal{D}_{\mathrm{val}}$}
    \State $\mathcal{U}_A^{(t)},\mathcal{U}_B^{(t)} \leftarrow \textsc{RandomPartition}(\mathcal{U},p=0.5)$
    \State Clear gradients

    \State Compute $\mathcal{L}_A^{(t)}$ with modules in $\mathcal{U}_A^{(t)}$ disabled
    \State Backpropagate $\mathcal{L}_A^{(t)}$ and accumulate gradients

    \State Compute $\mathcal{L}_B^{(t)}$ with modules in $\mathcal{U}_B^{(t)}$ disabled
    \State Backpropagate $\mathcal{L}_B^{(t)}$ and accumulate gradients

    \For{each $(l,m)\in\mathcal{U}$}
        \State $
        g_{l,m}^{(t)}
        \leftarrow
        \nabla_{e^{l,m}}
        \left(
        \mathcal{L}_A^{(t)}
        +
        \mathcal{L}_B^{(t)}
        \right)
        $
        \State $
        s_{l,m}^{(t)}
        \leftarrow
        \frac{1}{r}
        \left\|
        e^{l,m}
        \odot
        g_{l,m}^{(t)}
        \right\|_1
        $
        \State Append $s_{l,m}^{(t)}$ to $\mathcal{S}_{l,m}$
    \EndFor
\EndFor

\For{each $(l,m)\in\mathcal{U}$}
    \State $
    H(l,m)
    \leftarrow
    \operatorname{Mean}(\mathcal{S}_{l,m})
    \cdot
    \operatorname{Std}(\mathcal{S}_{l,m})
    $
\EndFor

\State \Return $H$
\end{algorithmic}
\end{algorithm}

\begin{algorithm}[t]
\caption{Hybrid-LoRA Training}
\label{alg:hybrid_lora_overview}
\begin{algorithmic}[1]
\Require Pretrained LLM $\mathcal{M}_0$, candidate module set $\mathcal{U}=\mathcal{A}\times\{1,\dots,L\}$, warmup steps $T_{\mathrm{warm}}$, validation dataloader $\mathcal{D}_{\mathrm{val}}$, FFT budget ratio $R_{\mathrm{fft}}$
\Ensure Hybrid fine-tuned model $\mathcal{M}_{\mathrm{hybrid}}$

\State Initialize probing model $\mathcal{M}_{\mathrm{probe}} \leftarrow \mathcal{M}_0$
\State Attach LoRA modules to all candidate modules $(l,m)\in\mathcal{U}$ on probing model $\mathcal{M}_{\mathrm{probe}}$
\State Freeze pretrained weights and train only LoRA parameters
\State Warm up $\mathcal{M}_{\mathrm{probe}}$ for $T_{\mathrm{warm}}$ steps

\State Compute Hybrid-Scores for all modules:
\[
H(l,m) \leftarrow \textsc{HybridScore}(\mathcal{M}_{\mathrm{probe}}, \mathcal{U}, \mathcal{D}_{\mathrm{val}})
\]

\State Rank all modules in $\mathcal{U}$ by $H(l,m)$ in ascending order
\State Select top-ranked modules under the FFT budget $R_{\mathrm{fft}}$ as $S_{\mathrm{fft}}$
\State Define the remaining modules as
\[
S_{\mathrm{LoRA}} = \mathcal{U}\setminus S_{\mathrm{fft}}
\]

\State Reinitialize hybrid model $\mathcal{M}_{\mathrm{hybrid}} \leftarrow \mathcal{M}_0$
\State Unfreeze full parameters of modules in $S_{\mathrm{fft}}$
\State Attach LoRA modules to modules in $S_{\mathrm{LoRA}}$
\State Train $\mathcal{M}_{\mathrm{hybrid}}$ with the formal post-training objective

\State \Return $\mathcal{M}_{\mathrm{hybrid}}$
\end{algorithmic}
\end{algorithm}

\clearpage

\end{document}